\documentclass[journal,comsoc]{IEEEtran}

\usepackage{booktabs} 
\usepackage{caption}
\usepackage{subcaption}
\usepackage{multirow}
\usepackage{longtable}
\usepackage{listings}
\usepackage{xcolor,colortbl}
\definecolor{maroon}{cmyk}{0,0.87,0.68,0.32}
\definecolor{orange}{rgb}{1,0.5,0}
\definecolor{green}{rgb}{70,130,180}
\usepackage{dirtytalk}
\usepackage{tikz}

\usepackage{array}
\usepackage[labelfont=bf]{caption}
\usepackage{xcolor} \usepackage{multirow, ltablex, colortbl, makecell}

\usepackage{ragged2e}
\usepackage{lipsum}
\usepackage{amssymb}
\usepackage{pifont}
\ifCLASSOPTIONcompsoc
  \usepackage[nocompress]{cite}
\else
  \usepackage{cite}
\fi
\ifCLASSINFOpdf
  
\fi
\usepackage{amsmath}
\usepackage{comment}

\usepackage{algorithmicx}
\usepackage[ruled,vlined]{algorithm2e}

\begin{document}

\title{Adversarial Attacks on Machine Learning  \\ Cybersecurity Defences in Industrial Control Systems}
%
%
%
%

\author{Eirini Anthi, Lowri Williams, Matilda Rhode, Pete Burnap, Adam Wedgbury} 

\author{
Eirini Anthi\textsuperscript{1,*}\thanks{*Corresponding author: anthies@cardiff.ac.uk}, 
Lowri Williams\textsuperscript{1}, Matilda Rhode\textsuperscript{1}, 
Pete Burnap\textsuperscript{1}, Adam Wedgbury\textsuperscript{2} 
\\ \textsuperscript{1}Cardiff University, School of Computer Science \& Informatics, Cardiff, UK 
\\ \textsuperscript{2}Airbus, Newport, UK}

\IEEEtitleabstractindextext{%
\begin{abstract}

The proliferation and application of machine learning based Intrusion Detection Systems (IDS) have allowed for more flexibility and efficiency in the automated detection of cyber attacks in Industrial Control Systems (ICS). However, the introduction of such IDSs has also created an additional attack vector; the learning models may also be subject to cyber attacks, otherwise referred to as Adversarial Machine Learning (AML). Such attacks may have severe consequences in ICS systems, as adversaries could potentially bypass the IDS. This could lead to delayed attack detection which may result in infrastructure damages, financial loss, and even loss of life. This paper explores how adversarial learning can be used to target supervised models by generating adversarial samples using the Jacobian-based Saliency Map attack and exploring classification behaviours. The analysis also includes the exploration of how such samples can support the robustness of supervised models using adversarial training. An authentic power system dataset was used to support the experiments presented herein. Overall, the classification performance of two widely used classifiers, Random Forest and J48, decreased by 16 and 20 percentage points when adversarial samples were present. Their performances improved following adversarial training, demonstrating their robustness towards such attacks.

\end{abstract}

}

\maketitle

\IEEEdisplaynontitleabstractindextext

%
\IEEEpeerreviewmaketitle

\ifCLASSOPTIONcompsoc
\IEEEraisesectionheading{\section{Introduction}\label{sec:introduction}}
\else
\section{Introduction}
\label{sec:introduction}
\fi

\IEEEPARstart{I}{ndustrial} Control Systems (ICS) play a key role in Critical National Infrastructure (CNI) concepts such as manufacturing, power/smart grids, water treatment plants, gas and oil refineries, and health-care. Historically, ICS networks and their components were protected from cyber attacks as they ran on proprietary hardware and software, and were connected in isolated networks with no external connection to the Internet \cite{kravchik2018detecting}. However, as the world is becoming more interconnected, there has been a need to connect ICS components together and to other networks, allowing remote access and monitoring functionalities. As a result, ICS are now subject to a range of security vulnerabilities \cite{kravchik2018detecting}. 

Given the importance of these systems, they have become an attractive target to an attacker. As these systems control operations in the physical world, the cyber attacks against them may have major consequences for the environment they operate in, and subsequently, its users. It is therefore understandable that the security issues surrounding such systems have become a global issue. Thus, designing robust, secure, and efficient mechanisms for detecting and defending cyber attacks in ICS networks is more important than ever \cite{ashibani2017cyber}. 
 
Although there exist several security mechanisms for traditional IT systems, their integration into ICS systems is challenging mainly for two reasons; a) ICS devices are resource constrained, and b) they include legacy systems and devices that do not support modern security measures. Subsequently, complementary security solutions, such as passive process data monitoring, are promising \cite{erba2019real}. This has led to a substantial increase in research focusing on ICS tailored Intrusion Detection Systems (ICS). Such intrusion systems operate by observing the network or sensor data in order to detect attacks and anomalies that may affect ICS. 

Due to their efficiency in detecting attacks, there has been a substantial increase in the application and integration of machine learning within IDSs  (e.g.\cite{beaver2013evaluation, bigham2003safeguarding, teixeira2018scada, strauss2017ensemble, goh2017anomaly, maglaras2014intrusion,kravchik2018detecting, linda2009neural}). However, the introduction of such systems has introduced an additional attack vector; the trained models may also be subject to attacks. The act of deploying attacks towards machine learning based systems is known as Adversarial Machine Learning (AML). The aim is to exploit the weaknesses of the pre-trained model which has ``blind spots" between data points it has seen during training. More specifically, by automatically introducing slight perturbations to the unseen data points the model may cross a decision boundary and classify the data as a different class. As a result, the model's effectiveness can be reduced as it is presented with unseen data points that it cannot associate target values to, subsequently increasing the number of misclassifications. 

The existence of such techniques means that infrastructures which incorporate machine learning based IDSs may be at risk of being vulnerable to cyber attacks. In the context of ICS, AML can be used to manipulate data from actuators or other devices by including perturbations to cause malicious data to be classified as being benign, consequently bypassing the IDS. This could lead to delayed attack detection, information leakage, financial loss, and even loss of life. It is therefore understandable that as machine learning based detection mechanisms become more widely deployed, the adversary incentive for defeating them increases. As a result, it is evident that machine learning based IDSs must be extensively evaluated against AML attacks.

To the best of our knowledge, this is the first study which investigates the behaviour of supervised models against AML, as well as the defence of such attacks in the context of ICS. The main contributions of the work presented in this paper are the empirical investigations into:

\begin{itemize}
    \item generating adversarial samples from a power system dataset
    \item the behaviour of supervised machine learning algorithms against adversarial samples for intrusion detection in an ICS system
    \item how adversarial training can support the robustness of such models
\end{itemize}

\begin{figure}[ht]
  \centering
  \vspace{3mm} 
  \includegraphics[width=0.3\textwidth]{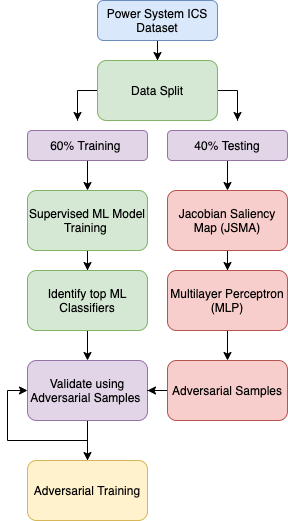}
\caption{An overview of the study design}
\label{study_design}
\end{figure}

The study uses a representative power system dataset and was designed as follows (see Figure \ref{study_design}): 1) randomly split the power system dataset into training and testing set, each containing 60\% and 40\% data points respectively, 2) evaluate a range of supervised machine learning models and identify which are the best performing, 3) generate adversarial samples using the Jacobian-based Saliency map method, 4) evaluate the performance of the trained models in 2 on the generated adversarial samples in 3, 5) include a percentage of adversarial samples from 3 in the training data and re-train and evaluate the models.

The remainder of this paper is structured as follows: Section \ref{sec:background} discusses the relevant work in this research area, Section \ref{sec:icscase} discusses the power system testbed and the generated dataset which is used to support the experiments in this paper, Section \ref{sec:supervisedlearning} evaluates the performance of a range of supervised classifiers, Section \ref{sec:aml} discusses AML and the methodology followed to generate adversarial samples, Section \ref{sec:defending} investigates the effectiveness of adversarial training as a defence mechanisms, and finally \ref{sec:conclusion} concludes the paper.

\section{Related Work}\label{sec:background}

There has been a substantial increase in machine learning based IDSs for a range of ICS systems. Table \ref{tab:ics} presents a summary of the existing ICS systems and associated supervised learning approaches to attack detection and classification in these contexts. To date, there has been less focus on AML in this context. Such research has mainly focused on email spam classifiers, malware detection, and very recently there has been interest in AML against network IDSs for traditional networks (e.g. \cite{rigaki2017adversarial, biggio2010multiple, huang2011adversarial}). 

More specifically, both Nelson et al. \cite{nelson2008exploiting} and Zhou et al. \cite{zhou2012adversarial} demonstrate that an adversary can exploit and successfully bypass the machine learning methods employed in spam filters by modifying a small percentage of the original training data. Moreover, Grosse et al. \cite{grosse2017adversarial} evaluate the robustness of a neural network trained on the DREBIN Android malware dataset. They report that it is possible to confuse the model by perturbing a small amount of the features in the training set. Such an attack is considered to be a white box attack, as in order to be successful, the adversary needs to have access or knowledge of the dataset and the features it includes. Furthermore, Hu and Tan \cite{hu2017generating} proposed a more advanced adversarial technique which uses the concept of Generative Adversarial Networks (GAN) to successfully attack malware classifiers without requiring any knowledge of the data and the system. This is known as a black box attack.

In the context of ICSs, there exist only a handful of investigations into AML attacks. Specifically, Zizzo et al. \cite{zizzo2019adversarial} showcased a simple AML attack against an Long Short-Term Memory (LSTM) classifier which was applied on an ICS dataset. However, this work is at a preliminary stage as the adversarial samples were generated by manually selecting the feature/acutator values to be perturbed. Yaghoubi and Fainekos \cite{yaghoubi2019gray} proposed a gradient based search approach which was evaluated on a Simulink model of a steam condenser. However, this approach is efficient only against a handful of systems that may specifically employ Recurrent Neural Networks (RNN) with smooth activation functions. Finally, Erba et al. \cite{erba2019real} demonstrated two types of real-time evasion attacks, again using Recurrent Neural Network models, and used an autoencoder to generate adversarial samples, which is computationally complex. Neither of these aforementioned works investigate \textit{defence methods against AML}. Conclusively, it is evident that there is room to investigate AML and the defence against such attacks for current IDSs in ICS systems that are supported by supervised learning. Moreover, as Table \ref{tab:ics} shows, Recurrent Neural Networks are yet to gain prominence in attack detection in an ICS context - with algorithms such as Naive Bayes, Random Forest, SVM, and J48 being much more widely used. We therefore base our experiments in defending against AML on these methods as the state of the art in ML-driven attack detection methods for ICS.

\begin{table*}[ht]
\centering
\begin{tabular}{|c|c|l|}
\hline
\textbf{Work} & \textbf{Dataset} & \textbf{Machine Learning Models} \\ \hline
\cite{beaver2013evaluation} & Gas Pipeline & Naive Bayes, Random Forest, SVM, J48, OneR \\ \hline
\cite{morris2015industrial} & Power System & OneR, Random Forest, Naive Bayes, SVM, JRipper + Adaboost \\ \hline
\cite{anton2018evaluation} & Power system (synthetic) & Naive Bayes, Random Forests, SVM \\ \hline
\cite{robles2018supervised} & SWaT & SVM, J48, Random Forest \\ \hline
\cite{ullah2017hybrid} & SCADA/ICS & J48, Naive Bayes \\ \hline
\cite{perez2018machine} & Gas Pipeline & SVM, Random Forest \\ \hline
\cite{qu2017instruction} & SCADA/Modbus & Decision Tree, K-Nearest Neighbor, SVM, OCSVM \\ \hline
\cite{teixeira2018scada} & SCADA Testbed & Random Forest, J48, Logistic Regression, Naive Bayes \\ \hline
\cite{yeckle2017evaluation} & Power Grid, Water Plant, Gas Plant & J48, Random Forest, Naive Bayes, SVM, JRipper + Adaboost \\ \hline
\cite{hoxha2019supervised} & Wind Turbines & SVM \\ \hline
\cite{frazao2018denial} & SCADA Testbed & SVM, Decision Tree, and Random Forest \\ \hline
\cite{lahza2018applying} & Power System & SVM, J48, Neural Network \\ \hline
\cite{werling2014behavioral} & SCADA & Naive Bayes, BayesNet, J48 \\ \hline
\cite{siddavatam2017ensemble} & SCADA Testbed & Decision Tree, Random Forest \\ \hline
\cite{wang2019detection} & Power System & Random Forest \\ \hline
\cite{abdallah2018fault} & Wind Turbine & Decision Trees (J48, Random Forest, CART, Ripper, etc.) \\ \hline
\cite{bigham2003safeguarding} & SCADA Testbed & Bayesian Network \\ \hline
\cite{gao2019lstm} & SCADA Testbed & Long Short Term Memory (RNN)  \\ \hline
\cite{kravchik2018detecting} & SWaT & 1D Convolutional Networks \\ \hline
\cite{linda2009neural} & ICS Testbed & Neural Network (Error-back propagation and Levenberg-Marquardt) \\ \hline
\cite{goh2017anomaly} & SWaT & Long Short Term Memory (RNN) \\ \hline
\cite{maglaras2014intrusion} & SCADA Network Traffic & One-Class SVM \\ \hline
\cite{feng2017multi} & ICS Testbed & Long Short Term Memory (RNN)  \\ \hline
\end{tabular}
\caption{Summary of current work on Intrusion Detection Systems in Industrial Control Systems}
\label{tab:ics}
\end{table*}

\section{Industrial Control System Case Study: Power System }\label{sec:icscase}

\begin{figure}[ht]
  \centering
  \vspace{3mm} 
  \includegraphics[width=0.45\textwidth]{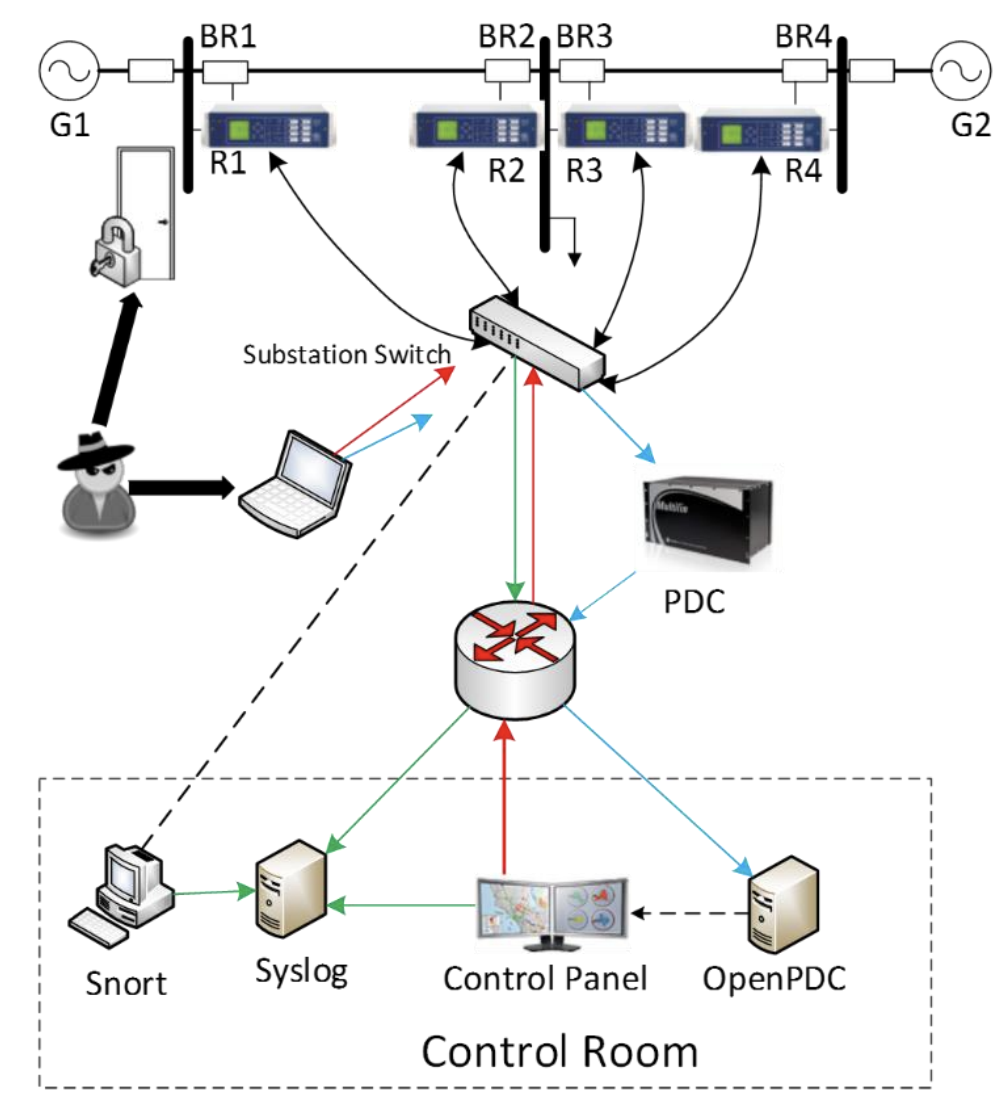}
\caption{Power System Framework Testbed \cite{PowerSys58:online}}
\label{powersystem}
\end{figure}

Mississippi State University and Oak Ridge National Laboratory implemented a scaled-down version of a power system framework. Although this system is relatively small, it captures the core function and is considered as being a representative example of a larger power system \cite{pan2015classification}. Figure \ref{powersystem} illustrates in more detail the power system framework configuration and the components used for generating the datasets in which support the experiments in this paper. 

More specifically, the components of the power system include:

\begin{itemize}
    \item G1 and G2 are the main generators.
    \item R1, R2, R3, and R4 are the Intelligent Electronic Devices (IEDs) responsible for switching the breakers (BR1, BR2, BR3, BR4), which are automatically operated electrical switches designed to protect electrical circuits from damage caused by excess current from an overload or short circuit, on and off.   
    \item  Each IED automatically controls one breaker (e.g. R1 controls BR1, R2 controls BR2, etc.)
    \item The IEDs use a distance protection scheme which trips the breaker on detected faults (whether they are valid or invalid) since they have no internal validation to detect the difference.
    \item Operators can also manually issue commands to the IEDs to manually trip the breakers. The manual override is used when performing maintenance on the lines or other system components.
    \item There are also other network monitoring devices connected on the testbed, such as SNORT and Syslog servers.
\end{itemize}

\subsection{Dataset}\label{dataset}

A total of 15 datasets containing both benign and malicious data points were generated from the power system testbed by \cite{pan2015classification}. These data points have been further categorised into three main classes; `no event' instances, `natural event' instances, and `attack event' instances. Both the `no event' and `natural event' instances are grouped together to represent benign activity. To generate the malicious data, attacks from 5 scenarios were deployed on the power system. These attacks are described as follows:

\begin{enumerate}
    \item \textbf{Short-circuit fault}. This is a short in a power line and can occur in various locations along the line. The location is indicated by the percentage range.
    \item \textbf{Line maintenance}. One or more relays are disabled on a specific line to do maintenance for that line.
    \item \textbf{Remote tripping command injection attack}. This is an attack that sends a command to a relay which causes a breaker to open. It can only be done once an attacker has penetrated outside defenses.
    \item \textbf{Relay setting change attack}. Relays are configured with a distance protection scheme. The attacker changes the setting to disable the relay function so that the relay will not trip for a valid fault or a valid command.
    \item \textbf{Data injection attack}. A valid fault is imitated by changing values to parameters such as the current, voltage, and sequence components. This attack aims to blind the operator and causes a black out.
    
\end{enumerate}

For the purposes of the work described in this paper, all 15 datasets were used. The dataset consisted of 55,663 malicious and 22,714 benign data points.

\section{Supervised Machine Learning}\label{sec:supervisedlearning}

To explore how well supervised classification algorithms can learn to detect cyber attacks in an ICS environment, the performance of supervised machine learning when the corresponding data discussed in Section \ref{dataset} was used to train the classification model and evaluated. The following Sections report the features present in the power systems dataset, as well as describing the methodology behind selecting and training the best performing supervised classifiers.

\subsection{Feature Selection} \label{features}

In order to perform machine learning classification experiments, it is essential to identify which attributes best describe the dataset. In this case, the data points within the power system dataset contain attributes associated with synchrophasor measurements and basic network security mechanisms. A synchrophasor measurement unit is a device which measures the electrical waves on an electricity grid, using a common time source for synchronization. The dataset contains a total of 128 features \cite{PowerSys58:online}. These features are described in more detail as follows:

\begin{itemize}
    \item 29 types of measurements from each synchrophasor measurement unit. In this specific power system testbed, there are 4 PMUs. Therefore, the dataset contains a total of 116 synchrophasor measurement columns. 
    \item 12 types of measurements of control panel logs, snort alerts, and relay logs of the 4 synchrophasor measurement unit and relay.
    
\end{itemize}

Table \ref{features} summarises the features included in the dataset, as well as their corresponding descriptions. More specifically, the index of each feature is in the form of ``R\#-Signal Reference''. The ``R `\#' '' specifies the type of measurement from the synchrophasor measurement unit. For instance, ``R1-PA1:VH'' corresponds to the ``Phase A voltage phase angle'' measured by ``PMU R1''.

\begin{table}[]
\begin{tabular}{|c|c|}
\hline
\textbf{Feature} & \textbf{Description} \\ \hline
PA1-PA3:VH & PA1:VH – PA3:VH Phase A \\ \hline
PM1: V -PM3:V & C Voltage Phase Angle \\ \hline
PA4:IH - PA6:IH & Phase A - C Current Phase Angle \\ \hline
PM4: I – PM6: I & Phase A - C Current Phase Magnitude \\ \hline
PA7:VH – PA9:VH & Pos. – Neg. – Zero Voltage Phase Angle \\ \hline
PM7: V – PM9: V & Pos. – Neg. – Zero Voltage Phase Magnitude \\ \hline
PA10:VH - PA12:VH & Pos. – Neg. – Zero Current Phase Angle \\ \hline
PM10: V - PM1 & Pos. – Neg. – Zero Current Phase Magnitude \\ \hline
F & Frequency for relays \\ \hline
DF & Frequency Delta (dF/dt) for relays \\ \hline
PA:Z & Appearance Impedance for relays \\ \hline
PA:ZH & Appearance Impedance Angle for relays \\ \hline
S & Status Flag for relays \\ \hline
\end{tabular}
\caption{Features included as part of the power system dataset}
\label{tab:features}
\end{table}

\subsection{Model Training}\label{modeltraining}

To explore how well supervised machine learning algorithms can detect cyber attacks in an ICS environment, the corresponding power system dataset was used to evaluate a range of state-of-the-art classifiers. In the case of identifying whether a datapoint is malicious or benign, classification is evaluated relative to the training dataset, producing four outputs:

\begin{itemize}
    \item {true positives (TP) - data points are predicted as being malicious, when they are indeed malicious.}
    \item {true negatives (TN) - data points are predicted as being benign, when they are indeed benign.}
    \item {false positives (FP) - data points are predicted as being malicious, when in fact, they are benign.}
    \item {false negatives (FN) - data points are predicted as being benign, when in fact, they are malicious.}
\end{itemize}

Subsequently, these output are used to evaluate the classification performance of the trained model using Precision (P), Recall (R), and F1-score (F). Such measures are calculated using the equations in Equation \ref{prf}.

\begin{equation} \label{prf}
    P = \frac{TP}{TP + FP}, \quad R = \frac{TP}{TP + FN}, \quad F = 2 \cdot \frac{P \cdot R}{P + R}
\end{equation}

The ``no free lunch'' theorem suggests that there is no universally best learning algorithm \cite{no-free-lunch-theorem}. In other words, the choice of an appropriate algorithm should be based on its performance for that particular problem and the properties of data that characterize the problem. In this case, a variety of classifiers distributed as part of Weka \cite{weka-website} were evaluated using 10-fold cross-validation using their default hyper-parameters. 

To conform to other comparable IDSs in ICS systems in Table \ref{tab:ics}, the classifiers were also selected based on their ability to support a high-dimensional feature space. The classifiers included:

\begin{itemize}
    \item Generative models that consider conditional dependencies in the dataset or assume conditional independence (e.g. Bayesian Network, Naive Bayes).
    \item Discriminative models that aim to maximise information gain or directly maps data to their respective classes without modeling any underlying probability or structure of the data (e.g. J48 Decision Tree, Support Vector Machine).
\end{itemize}

\begin{figure}[ht]
  \centering
  \vspace{3mm} 
  \includegraphics[width=0.45\textwidth]{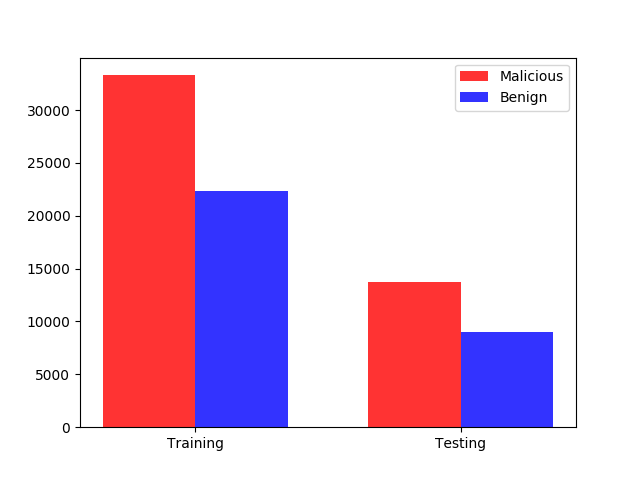}
\caption{Distribution of data points across both training and testing datasets}
\label{distribution_graph}
\end{figure}

To support classification experiments, a random subset of approximately 60\% of the dataset described in Section \ref{dataset} was selected for training, with the remaining 40\% selected for testing. Figure \ref{distribution_graph} reports the distributions of data points across the target values in both the training and testing datasets.

Table \ref{tab:crossvalidation} illustrates the results for each classifier. Previous work which have used a very small sample of this power system dataset to support their classification experiments have shown that the ensemble classifier which combines both the Adaboost and JRipper models was found to be the best performing \cite{hink2014machine}. However, in this work, when both the ensemble classifier and Support Vector Machine (SVM) classifier were applied, both models were still training following approximately 2 days of running. In this case, the models were stopped and their classification performances were omitted in the reporting of the results herein. Conversely, with F1-scores of 0.93 and 0.87, the classifiers with the highest performances were Random Forest and Weka's implementation of the J48 decision tree method with no pruning respectively.

\begin{table}[]
\centering
\begin{tabular}{|c|c|c|c|c|}
\hline
\textbf{Classifier} & \textbf{P} & \textbf{R} & \textbf{F} & \textbf{Time (s)} \\ \hline
Zero R & 0.50 & 0.70 & 0.58 & 0.20 \\ \hline
BayesNet & 0.66 & 0.61 & 0.63 & 8.28 \\ \hline
Naive Bayes & 0.67 & 0.31 & 0.19 & 54.00 \\ \hline
SVM & - & - & - & 28,800.00 \\ \hline
Adaboost + JRip & - & - & - & 28,800.00 \\ \hline
Random Forest & 0.94 & 0.93 & 0.93 & 247.00 \\ \hline
J48 & 0.87 & 0.87 & 0.87 & 480.00 \\ \hline
\end{tabular}
\caption{Weighted average results following cross-validation}
\label{tab:crossvalidation}
\end{table}

\section{Adversarial Machine Learning} \label{sec:aml}

To reiterate, the aim of AML is to automatically introduce perturbations to the unseen data points in order to confuse the pre-trained model. The following sections introduce the types of AML attacks, as well as the methods used to automatically generate adversarial samples.

\subsection{Adversarial Attack Types} 

Depending on the phase and aspect of the machine learning model that is being targeted, AML attacks can be described in terms of four primary vectors: \cite{barreno2006can, huang2011adversarial}:

\begin{itemize}

  \item The \textbf{Influence} of an attack's affects the classifier's decision. Attacks can be further categorised as causative attacks, which occur during the learning phase (poison attacks), or exploratory attacks, which target the trained model during the testing phases (evasion attacks). 
  
  \item \textbf{Security Violations} affect either the integrity of the model when the adversarial samples cause misclassifications, or when the high rate of misclassifications causes the model to become unusable. 
  
  \item \textbf{Specificity} refers to targeted attacks, where the adversarial samples aim to target a specific target value, or indiscriminate attacks, where the samples do not target a specific target value.
  
  \item \textbf{Privacy} refers to attacks where the adversary's goal is to extract information from the classifier. 
  
\end{itemize}

Papernot et al. \cite{papernot2016limitations} further categorise adversarial attacks based on:

\begin{itemize}

    \item Their \textbf{complexity}. The consequences of such attacks can range from slightly reducing the confidence of a model's predictions to causing it to misclassify all unseen data points.
    
    \item The \textbf{knowledge} an adversary may have. A \textit{white box} attack refers to when an attacker has useful knowledge related to the learning model, such as it's architecture, the network's traffic it reads, and the features used to support its training. It is considered as being a \textit{black box} attack when an adversary has  no information about the internal workings of the target model.
    
\end{itemize}

Given that we have access to the full training dataset and its features, and we don't know the target model,  the AML approach presented in this work is classified as a being a grey-box attack.

\subsection{Adversarial Sample Generation Methods}\label{adversarialmethods}

There are various methods by which adversarial samples can be generated. Such methods vary in complexity, the speed of their generation, and their performance. An unsophisticated approach towards crafting such samples is to manually perturb the input data points. However, manual perturbations are slow to generate and evaluate by comparison with automatic approaches. Two of the most popular techniques towards automatically generating perturbed samples include the Fast Gradient Sign Method (FGSM) and the Jacobian based Saliency Map Attack (JSMA), presented by Goodfellow et al. \cite{goodfellow2014generative} and Papernot et al. \cite{papernot2016limitations} respectively. 

Both methods rely on the methodology, that when adding small perturbations ($\delta$) to the original sample (X), the resulting sample (X*) can exhibit adversarial characteristics (X* = X +  $\delta$) \cite{rigaki2017adversarial} in that X* is now classified differently by the targeted model. Moreover, both methods are also usually applied by using a pre-trained MLP as the underlying model for the adversarial sample generation.

The FGSM method aims to target each of the features of the input data by adding a specified amount of perturbation. The perturbation noise is computed by the gradient of the cost function $J$ with respect to the input data. Let $\theta$ represent the model parameters, $x$ are the inputs to the model, $y$ are the labels associated with the input data, $\epsilon$ is a value which represents the extent of the noise to be applied, and $J$($\theta$,$x$,$y$) is the cost function used to train the targeted neural network.

\begin{equation*}x^{*}=x+\epsilon \operatorname{sign}\left(\nabla_{x} J(\theta, x, y)\right) \tag{1}\end{equation*}

On the other hand, the JSMA method generates perturbations using saliency maps. A saliency map identifies which features of the input data are the most relevant to the model decision being one class or another; these features if altered are most likely affect the classification of the target values. More specifically, an initial percentage of features ($\theta$) is chosen to be perturbed by a ($\gamma$) amount of noise. Then, the model establishes whether the added noise has caused the targeted model to misclassify or not. If the noise has not affected the model's performance, another set of features is selected and a new iteration occurs until a saliency map appears which can be used to generate an adversarial sample. 

Given that the JSMA method may take a few iterations to generate adversarial samples, the FGSM is computationally faster \cite{papernot2016limitations}. However, as opposed to FGSM which alters each feature, JSMA is a more complex and elaborate approach which represents more realistic attacks as it progressively alters a small percentage of features at a time. This allows for more realistic and finer grained AML attacks, as adversaries are able to define both the percentage of features to perturb and the amount of perturbation to include when generating the adversarial samples.

This work presents the use of JSMA in a grey-box attack, in which the attacker has no knowledge of the target model but has access to the full dataset and knowledge of features. Despite not knowing the target model, we can approximate samples that will cause the target model to misclassify using another model due to the transferability of adversarial samples across machine learning models \cite{papernot2016transferability}.  

In this case, the adversarial samples used in the experiments herein were generated using the JSMA method. A pre-trained MLP was used as the underlying model for the generation. The code implementation used to create the adversarial data was based on the CleverHans project \cite{papernot2016limitations}. Table \ref{adversarial_example} shows the transformation of the features of a malicious data point using the JSMA method.

\begin{table}[]
\centering
\begin{tabular}{|c|c|c|c|c|}
\hline
\textbf{Dataset}                & \textbf{R1-PA1:VH} & \textbf{R2:DF} & \textbf{R2-PM11:I} & \textbf{R3-PM5:I} \\ \hline
Original test data           & 0.764515           & 0.361399       & 0.008482           & 0.026826          \\ \hline
$\theta$ = 0.1, $\gamma$ = 0.1  & 0.765000              & 0.361000          & 0.008480            & 0.026800            \\ \hline
$\theta$ = 0.9,  $\gamma$ = 0.9 & 1.000000                  & 0.538000          & 0.008600             & 0.026800            \\ \hline
\end{tabular}
\caption{An example of how features are perturbed using JSMA}
\label{adversarial_example}
\end{table}

\section{Evaluating Supervised Models on Adversarial Samples} \label{evaluation}

Both the trained Random Forest and J48 models presented in Section \ref{modeltraining} were first evaluated against the original testing dataset. The F1-scores achieved by both classifiers were 0.67 and 0.66 respectively. The confusion matrix in Table \ref{confusion-matrix} shows how the predicted classes for each data point in the original testing dataset compare against the actual ones. In comparison to the Random Forest model, J48 demonstrated a high percentage of correct predictions, thus less often miclassifying the data points.

\begin{table}[]
\centering
\begin{tabular}{ccccccccc}
                                             &                                 & \multicolumn{2}{c}{Predicted}                                     &                      &                                              &                                 & \multicolumn{2}{c}{Predicted}                                     \\ \cline{3-4} \cline{8-9} 
                                             & \multicolumn{1}{c|}{}           & \multicolumn{1}{c|}{\textbf{0}} & \multicolumn{1}{c|}{\textbf{1}} &                      &                                              & \multicolumn{1}{c|}{}           & \multicolumn{1}{c|}{\textbf{0}} & \multicolumn{1}{c|}{\textbf{1}} \\ \cline{2-4} \cline{7-9} 
\multicolumn{1}{c|}{\multirow{2}{*}{Actual}} & \multicolumn{1}{c|}{\textbf{0}} & \multicolumn{1}{c|}{2,840}      & \multicolumn{1}{c|}{6,149}      &                      & \multicolumn{1}{c|}{\multirow{2}{*}{Actual}} & \multicolumn{1}{c|}{\textbf{0}} & \multicolumn{1}{c|}{10,610}     & \multicolumn{1}{c|}{3,115}      \\ \cline{2-4} \cline{7-9} 
\multicolumn{1}{c|}{}                        & \multicolumn{1}{c|}{\textbf{1}} & \multicolumn{1}{c|}{1,240}      & \multicolumn{1}{c|}{21,122}     &                      & \multicolumn{1}{c|}{}                        & \multicolumn{1}{c|}{\textbf{1}} & \multicolumn{1}{c|}{2,631}      & \multicolumn{1}{c|}{30,670}     \\ \cline{2-4} \cline{7-9} 
\multicolumn{1}{l}{}                         & \multicolumn{1}{l}{}            & \multicolumn{1}{l}{}            & \multicolumn{1}{l}{}            & \multicolumn{1}{l}{} & \multicolumn{1}{l}{}                         & \multicolumn{1}{l}{}            & \multicolumn{1}{l}{}            & \multicolumn{1}{l}{}            \\
                                             & \multicolumn{3}{c}{Random Forest}                                                                   &                      &                                              & \multicolumn{3}{c}{J48}                                                                            
\end{tabular}
\caption{Confusion matrices for the original test set (Benign = 0, Malicious = 1)}
\label{confusion-matrix}
\end{table}

\begin{figure}[ht]
  \centering
  \vspace{3mm} 
  \includegraphics[width=0.5\textwidth]{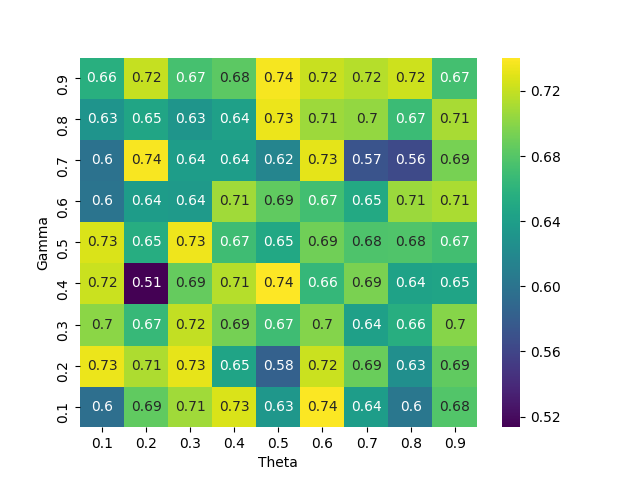}
\caption{Random Forest classification performance (F1-score) on adversarial samples generated using JSMA}
\label{heatmap-random}
\end{figure}

\begin{figure}[ht]
  \centering
  \vspace{3mm} 
  \includegraphics[width=0.5\textwidth]{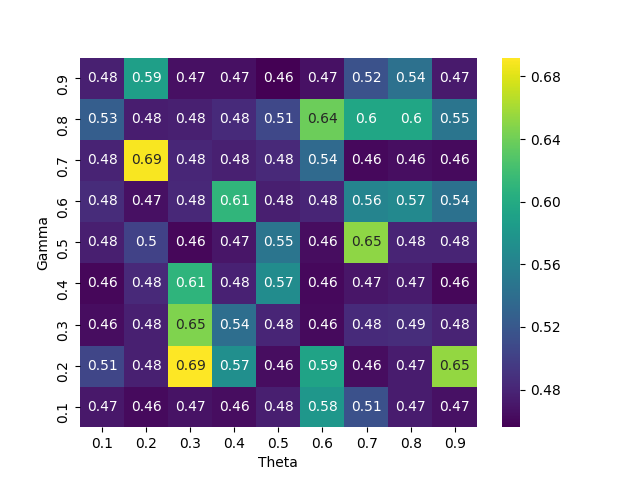}
\caption{J48 classification performance (F1-score) on adversarial samples generated using JSMA}
\label{heatmap-j48}
\end{figure}

To explore how different combinations of the JSMA parameters affect the performance of the trained classifiers, adversarial samples were generated from all malicious data points present in the testing data by using a range of combinations of $\theta$ and $\gamma$. The adversarial samples were joined with the benign testing data points and subsequently presented to the trained models. Figures \ref{heatmap-random} and \ref{heatmap-j48} report the overall weighted-averaged F1-scores for all adversarial combinations of JSMA's $\theta$ and $\gamma$ parameters. 

In comparison to Random Forest, the classification performance of the J48 model achieved a decrease in F1-scores across the majority of the $\theta$ and $\gamma$ parameters. This may indicate that J48 may be more sensitive, subsequently misclassifying malicious data points as benign. However, when $\theta$ = 0.3, $\gamma$ = 0.2 and $\theta$ = 0.2, $\gamma$ = 0.7, the model achieves a higher classification performance of 0.69 (an increase of 3 percentage points). This may indicate that the generation of some adversarial samples has made such data points more distinct in discriminating between the target values.   

Conversely, the classification performance of the Random Forest model achieved an increase in F1-scores for the majority of $\theta$ and $\gamma$ pairs. This may indicate that Random Forest may be a more robust classifier in correctly discriminating between malicious and benign data points. However, when $\theta$ = 0.2, $\gamma$ = 0.4, the model's classification performance decrease by 16 percentage points (F1-score = 0.57). Based on the dataset used in the experiments presented in this paper, $\theta$ = 0.2, $\gamma$ = 0.4 would be the optimal parameter an adversary would use to successfully reduce the accuracy of a machine learning based IDS, subsequently diverting malicious data points.  

These findings demonstrate the importance of parameter-tuning in applying JSMA for generating adversarial examples. The JSMA model is likely to be more robust under white-box conditions as it was designed but these results indicate that with careful parameter tuning, this approach can be adapted to work under black-box conditions. Although the F1-scores increase in some instances, the attacker is primarily interested in their malicious data points being classified as benign, such that an increase in F1-score is not necessarily undesirable from the attacker's perspective. 

The confusion matrices in Tables \ref{confusion-matrix-j48} and \ref{confusion-matrix-random} provide a better insight into the performance of the classifiers across the experiments. In comparison to the original classification distributions in Table \ref{confusion-matrix}, both classifiers demonstrate a significant increase in false positives. That is, data points with an actual target value of malicious have been misclassified as being benign. On the other hand, when $\theta$ = 0.5, $\gamma$ = 0.9, the Random Forest model's true positive distribution increases, which may explain as to why its F1-score also increases.

\begin{table}[]
\centering
\begin{tabular}{ccccccccc}
                                             &                                 & \multicolumn{2}{c}{Predicted}                                     &                      &                                              &                                 & \multicolumn{2}{c}{Predicted}                                     \\ \cline{3-4} \cline{8-9} 
                                             & \multicolumn{1}{c|}{}           & \multicolumn{1}{c|}{\textbf{0}} & \multicolumn{1}{c|}{\textbf{1}} &                      &                                              & \multicolumn{1}{c|}{}           & \multicolumn{1}{c|}{\textbf{0}} & \multicolumn{1}{c|}{\textbf{1}} \\ \cline{2-4} \cline{7-9} 
\multicolumn{1}{c|}{\multirow{2}{*}{Actual}} & \multicolumn{1}{c|}{\textbf{0}} & \multicolumn{1}{c|}{3,662}      & \multicolumn{1}{c|}{5,327}      &                      & \multicolumn{1}{c|}{\multirow{2}{*}{Actual}} & \multicolumn{1}{c|}{\textbf{0}} & \multicolumn{1}{c|}{3,662}      & \multicolumn{1}{c|}{5,327}      \\ \cline{2-4} \cline{7-9} 
\multicolumn{1}{c|}{}                        & \multicolumn{1}{c|}{\textbf{1}} & \multicolumn{1}{c|}{12,325}     & \multicolumn{1}{c|}{10,037}     &                      & \multicolumn{1}{c|}{}                        & \multicolumn{1}{c|}{\textbf{1}} & \multicolumn{1}{c|}{4,103}      & \multicolumn{1}{c|}{18,259}     \\ \cline{2-4} \cline{7-9} 
\multicolumn{1}{l}{}                         & \multicolumn{1}{l}{}            & \multicolumn{1}{l}{}            & \multicolumn{1}{l}{}            & \multicolumn{1}{l}{} & \multicolumn{1}{l}{}                         & \multicolumn{1}{l}{}            & \multicolumn{1}{l}{}            & \multicolumn{1}{l}{}            \\
                                             & \multicolumn{3}{c}{$\theta$ = 0.1 $\gamma$ = 0.3}                                                   &                      &                                              & \multicolumn{3}{c}{$\theta$ = 0.3 $\gamma$ = 0.2}                                                  
\end{tabular}
\caption{Confusion matrices after applying J48 to adversarial testing samples (Benign = 0, Malicious = 1)}
\label{confusion-matrix-j48}
\end{table}

\begin{table}[]
\centering
\begin{tabular}{ccccccccc}
                                             &                                 & \multicolumn{2}{c}{Predicted}                                     &                      &                                              &                                 & \multicolumn{2}{c}{Predicted}                                     \\ \cline{3-4} \cline{8-9} 
                                             & \multicolumn{1}{c|}{}           & \multicolumn{1}{c|}{\textbf{0}} & \multicolumn{1}{c|}{\textbf{1}} &                      &                                              & \multicolumn{1}{c|}{}           & \multicolumn{1}{c|}{\textbf{0}} & \multicolumn{1}{c|}{\textbf{1}} \\ \cline{2-4} \cline{7-9} 
\multicolumn{1}{c|}{\multirow{2}{*}{Actual}} & \multicolumn{1}{c|}{\textbf{0}} & \multicolumn{1}{c|}{2,840}      & \multicolumn{1}{c|}{6,149}      &                      & \multicolumn{1}{c|}{\multirow{2}{*}{Actual}} & \multicolumn{1}{c|}{\textbf{0}} & \multicolumn{1}{c|}{2,840}      & \multicolumn{1}{c|}{6,149}      \\ \cline{2-4} \cline{7-9} 
\multicolumn{1}{c|}{}                        & \multicolumn{1}{c|}{\textbf{1}} & \multicolumn{1}{c|}{9,732}      & \multicolumn{1}{c|}{12,630}     &                      & \multicolumn{1}{c|}{}                        & \multicolumn{1}{c|}{\textbf{1}} & \multicolumn{1}{c|}{1,024}      & \multicolumn{1}{c|}{21,338}     \\ \cline{2-4} \cline{7-9} 
\multicolumn{1}{l}{}                         & \multicolumn{1}{l}{}            & \multicolumn{1}{l}{}            & \multicolumn{1}{l}{}            & \multicolumn{1}{l}{} & \multicolumn{1}{l}{}                         & \multicolumn{1}{l}{}            & \multicolumn{1}{l}{}            & \multicolumn{1}{l}{}            \\
                                             & \multicolumn{3}{c}{$\theta$ = 0.2 $\gamma$ = 0.4}                                                   &                      &                                              & \multicolumn{3}{c}{$\theta$ = 0.5 $\gamma$ = 0.9}                                                  
\end{tabular}
\caption{Confusion matrices after applying Random Forest to adversarial testing samples (Benign = 0, Malicious = 1)}
\label{confusion-matrix-random}
\end{table}

\section{Defending Adversarial Machine Learning}\label{sec:defending}

A few methods towards defending AML attacks have been proposed in the literature. Two of the most popular techniques include adversarial training and adversarial sample detection. The former has been explored in the field of visual computing, where Goodfellow et al. \cite{goodfellow2014explaining} demonstrated that re-training the neural network on a dataset containing both the original and adversarial samples significantly improves its efficiency against adversarial samples. The latter technique involves the implementation of mechanisms that are capable of detecting the presence of such samples using direct classification, neural network uncertainty, or input processing \cite{zizzo2019adversarial}. However, these detection mechanisms have been found to be weak in defending AML \cite{athalye2018obfuscated, zizzo2019adversarial}.

Subsequently, in this paper, the robustness of supervised machine learning classifiers against AML is further evaluated using adversarial training. In this case, a random sample of 20\% of the adversarial data points in the testing dataset which significantly decreased the model's performance (Random Forest: $\theta$ = 0.2, $\gamma$ = 0.4 and J48: $\theta$ = 0.1, $\gamma$ = 0.3) were included in the original training dataset. 

The experiments described in Sections \ref{modeltraining} and \ref{evaluation} were repeated by retraining the models with the newly generated training data and applying such models on all unseen adversarial samples. Both the Random Forest and J48 models achieved cross-validation F1-scores of 0.94 and 0.89 respectively.

\begin{figure}[ht]
  \centering
  \vspace{3mm} 
  \includegraphics[width=0.5\textwidth]{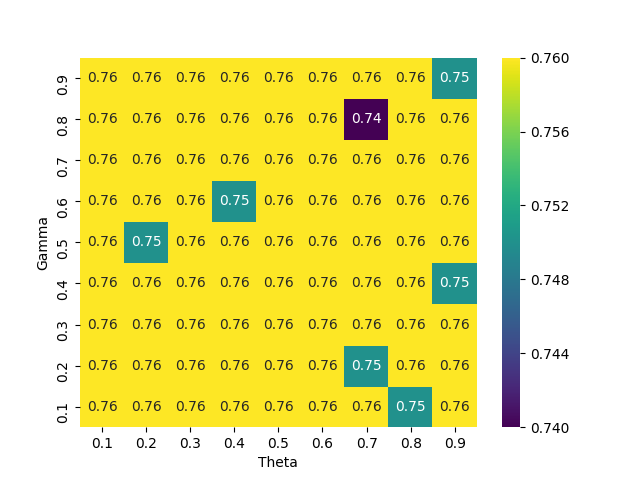}
\caption{Random Forest classification performance following adversarial training ($\theta$ = 0.2, $\gamma$ = 0.4)}
\label{heatmap-randomf-new}
\end{figure}

\begin{figure}[ht]
  \centering
  \vspace{3mm} 
  \includegraphics[width=0.5\textwidth]{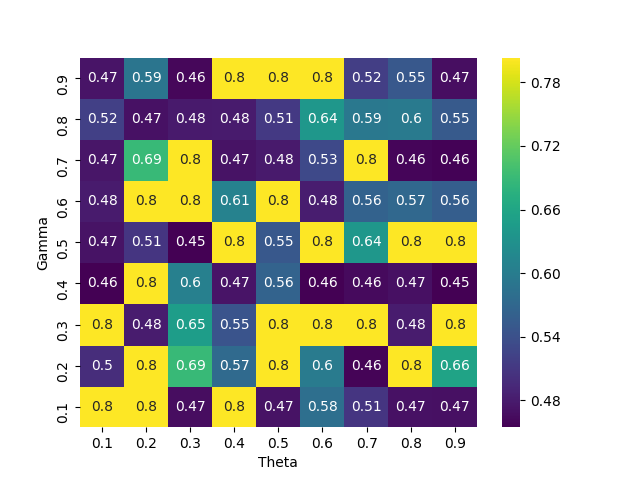}
\caption{J48 classification performance following adversarial training ($\theta$ = 0.1, $\gamma$ = 0.3)}
\label{heatmap-j48-new}
\end{figure}

Figures \ref{heatmap-randomf-new} and \ref{heatmap-j48-new} report the overall weighted-averaged F1-scores for all adversarial combinations of JSMA's $\theta$ and $\gamma$ parameters following adversarial training. The results demonstrated that for both classifiers, including adversarial samples in the training data increased their classification performances. More specifically, Random Forest and J48 achieved F1-scores of 0.76 and 0.80 respectively, an increase of 2 and 11 percentage points in comparison to the classification performances reported in Figures \ref{heatmap-random} and \ref{heatmap-j48}.

The classification performances demonstrated by the Random Forest model achieves a greater overall increase in comparison to the J48 model. That is, for Random Forest, the classification performance for all combinations were improved. Whereas for J48, only around 30\% of the classification performances increased significantly. This may imply that Random Forest is a more robust model towards classifying adversarial samples of all combinations of JSMA's $\theta$ and  $\gamma$ parameters. This is intuitive given Strauss et al.'s \cite{strauss2017ensemble} demonstration that ensemble machine learning algorithms are more robust against adversarial techniques and Random Forests are ensembles of decision trees (such as J48).    

\section{Conclusion} \label{sec:conclusion}

Due to their effectiveness and flexibility, machine learning based IDSs are now recognised as fundamental tools for detecting cyber attacks in ICS systems. Nevertheless, such systems are vulnerable to attacks that may severely undermine or mislead their capabilities, commonly known as Adversarial Machine Learning (AML). Such attacks may have severe consequences in ICS infrastructures, as adversaries could potentially modify malicious data points in order to bypass the IDS, causing delayed attack detection and extensive damages. Thus, it is evident that understanding the applicability of these attacks in ICS systems is necessary in order to develop more robust machine learning based IDSs.  

This paper explores how adversarial learning can be used to target supervised models by generating adversarial samples and exploring classification behaviours. To support the experiments presented herein, an authentic power system dataset was used to train and test widely used supervised machine learning classifiers. The testing data was presented to a JSMA in order to generate adversarial samples with a range of combinations that affect the amount of noise and the number of features to perturb. Such samples were evaluated against two of the best performing classifiers, Random Forest and J48. Overall, the classification performance for both models decreased by 16 and 20 percentage points when adversarial samples were present.

The analysis also includes the exploration of how such samples can support the robustness of supervised models using adversarial training. A random sample of 20\% of the generated adversarial data points were included in the original training dataset. The models were retrained and applied on all unseen adversarial samples. Overall, the classification performance of the Random Forest model reported a greater increase in comparison to the J48 model. This demonstrates that Random Forest is a more robust model towards classifying adversarial samples of all combinations of JSMA parameters on the given dataset.

\section{Future Work}

Although the experiments described in this paper have demonstrated that adversarial samples can successfully be generated using JSMA and affect the classification performance of state-of-the-art supervised models, it is important to note that there are several other methods of generating such samples to consider (e.g. Iterative Gradient Sign, Carlini Wagner, Generative Adversarial Networks). In this case, as part of future work, this study can be extended further to include different models as a source for generating adversarial samples. Moreover, AML should be further investigated against other models such as LSTMs. 

Finally, the robustness of the supervised models was demonstrated using adversarial training. It is also important to note that this method may not always be sufficient as it is difficult to anticipate all possible types of adversarial machine learning attacks against a given system. Therefore, there is a need to investigate other possible defense mechanisms.

\bibliographystyle{IEEEtran}
\bibliography{ref}

\end{document}